\pdfoutput=1

\documentclass[11pt]{article}

\usepackage{acl}

\usepackage{times}
\usepackage{latexsym}
\usepackage[T1]{fontenc} 
\usepackage[utf8]{inputenc}
\usepackage{microtype} 
\usepackage{inconsolata}
\usepackage{graphicx} 
\usepackage{enumitem}

\usepackage{times}
\usepackage{latexsym}
\usepackage{booktabs}
\usepackage{multirow} 
\usepackage{float}
\usepackage{graphicx}
\usepackage{capt-of} %
\usepackage{cuted}

\usepackage[T1]{fontenc}
\usepackage[utf8]{inputenc}
\usepackage{microtype}
\usepackage{inconsolata}
\usepackage{graphicx}
\usepackage{amsmath}
\usepackage{graphicx} %
\title{\raisebox{-0.5em}{\includegraphics[height=1.8em]{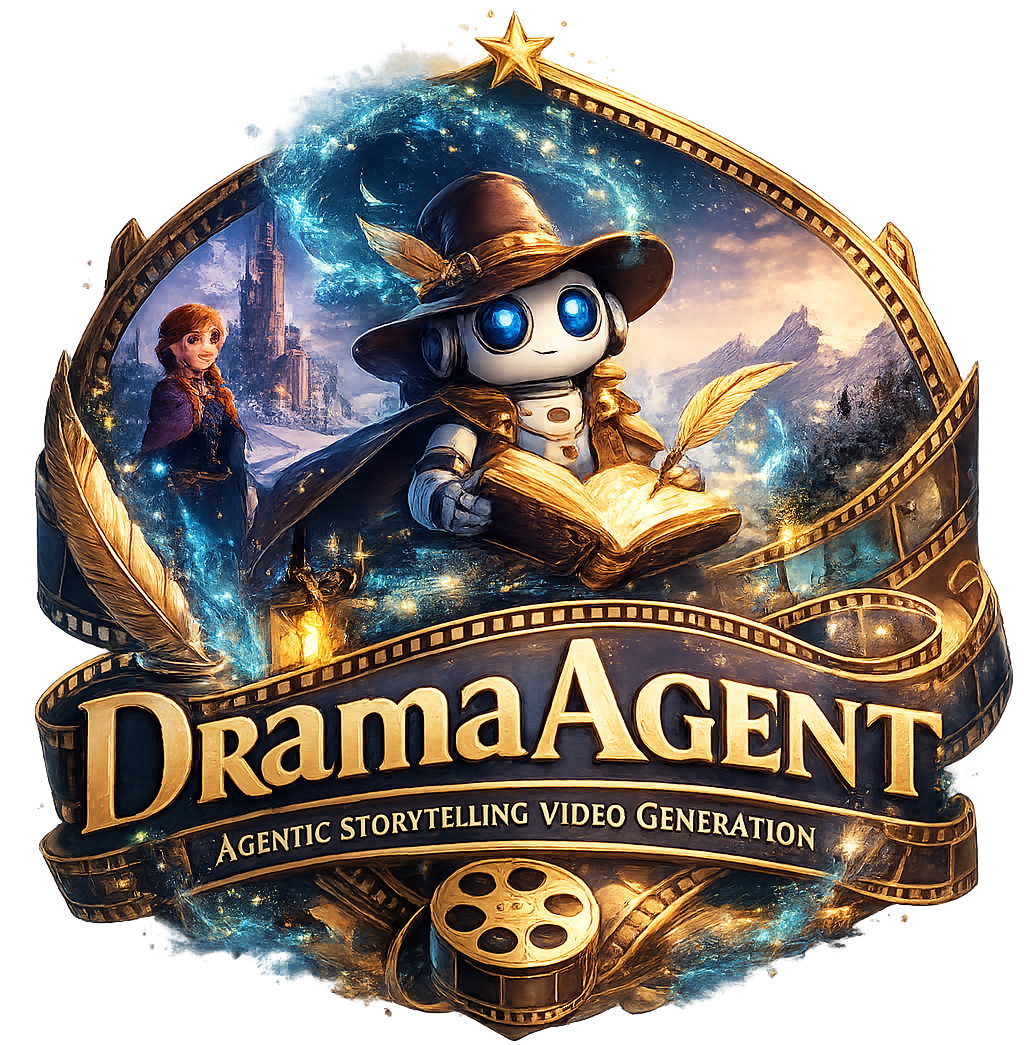}}~DramaAgent: Agentic Storytelling Video Generation}

\author{
  \textbf{Ting Huang$^{1*}$~~
  Biao Wu$^{2*}$~~
  Ronghao Chen$^{1*}$~~
  Zeyu Zhang$^{1\dag}$}\\ 
  \textbf{
  Tengfei Cheng$^{3}$~~
  Qizhen Lan$^{4}$~~
  Huacan Wang$^{5}$~~ 
  Hao Tang$^{1\ddag}$} \\
  $^1$Peking University~~
  $^2$UTS AAII~~
  $^3$UOL~~
  $^4$UTHealth Houston~~
  $^5$UCAS\\
  \small$^{*}$Equal contribution. $^{\dag}$Project lead. $^{\ddag}$Corresponding author: bjdxtanghao@gmail.com.
}

\begin{document}
\maketitle

\begin{strip}
\centering
\includegraphics[width=\textwidth]{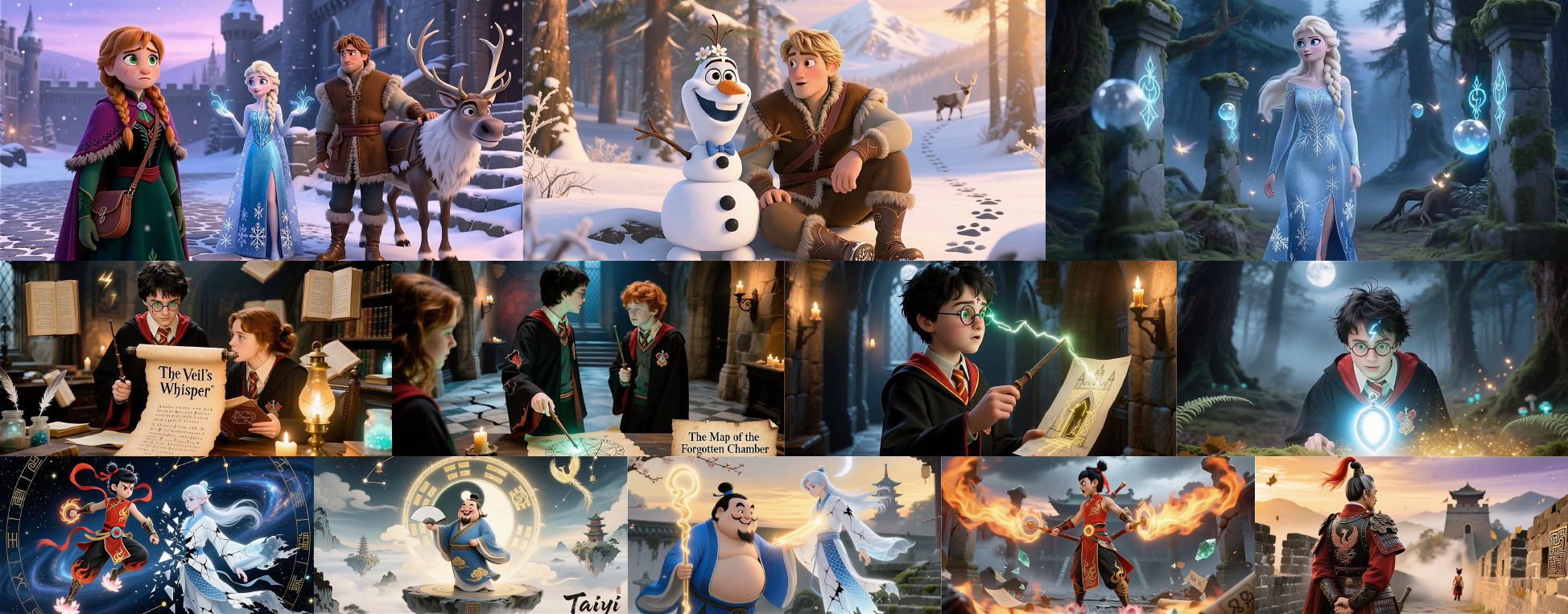}
\captionof{figure}{\textbf{DramaAgent} generates long-form animated stories from a narrative prompt via hierarchical planning, persistent character conditioning, and multimodal generation, enabling improved cross-scene coherence and story consistency.}
\label{fig:teaser}
\end{strip}

\begin{abstract}
Recent diffusion and autoregressive models have substantially improved text-to-video generation, yet producing coherent long-form story videos with consistent characters and aligned audio remains challenging.
Existing methods often suffer from narrative drift, unstable character identity, weak cross-scene continuity, and audio-visual mismatch over extended sequences.
We propose \textbf{DramaAgent}, a hierarchical, agentic, and model-agnostic framework for long-form text-to-video-and-audio generation.
Rather than improving the underlying video backbone itself, DramaAgent introduces an upper-level control layer that decomposes generation into story planning, persistent character conditioning, scene-wise synthesis, and reflection-guided targeted repair.
The framework maintains reusable story and character states across scenes, diagnoses failures such as identity drift, missing scene semantics, temporal discontinuity, and cross-modal mismatch, and repairs problematic clips in a stage-specific manner.
Experiments across multiple video generation backbones show that DramaAgent improves long-horizon coherence, character consistency, narrative fidelity, and scene-level audio-visual consistency over direct generation and strong baselines.
These results suggest that hierarchical agentic control is a practical direction for controllable long-form audiovisual generation.
Code: \url{https://github.com/AIGeeksGroup/DramaAgent}.
Website: \url{https://aigeeksgroup.github.io/DramaAgent}.
\end{abstract}

\section{Introduction}
Recent diffusion and autoregressive models have substantially advanced text-to-video generation, especially for short-form visual synthesis~\cite{videodiffusionmodel,videoworldsimulators2024,imagenvideo,zheng2024videogen,kong2024hunyuanvideo}.
However, generating coherent \emph{long-form story videos} remains difficult.
As generation spans multiple scenes, existing methods often suffer from narrative drift, unstable character identity, weak temporal continuity, and limited audio-visual consistency~\cite{wei2024dreamvideo,zheng2024videogen,zhou2022magicvideo,videodiffusionmodel,wu2023tune}.
These challenges are particularly severe in story-centric applications such as educational animation, narrative video creation, and marketing content, where coherence must be preserved over long horizons rather than within isolated short clips~\cite{polyak2024movie,wu2024moviebench,zhu2023moviefactory,hu2024storyagent,long2024videostudio}.

A straightforward solution is to use stronger video generators.
Yet even strong generators and commercial APIs are still primarily optimized for short clips, and their direct use for long-form storytelling remains brittle.
One-shot prompting tends to accumulate semantic errors over time, while simple clip stitching often fails to preserve consistent characters, scene transitions, and story state.
Meanwhile, fully end-to-end training for long-form video remains costly and difficult to adapt to rapidly evolving video backbone ecosystems.
This motivates a complementary perspective: instead of modifying the underlying generator, can long-form story generation be improved through a structured control framework built above heterogeneous generators?

We propose \textbf{DramaAgent}, a hierarchical, agentic, and model-agnostic framework for long-form text-to-video-and-audio generation.
DramaAgent treats long-form storytelling as a structured control problem over heterogeneous video and audio generators, rather than as independent short-clip synthesis.
It organizes generation into story planning, persistent character conditioning, scene-wise synthesis, and reflection-guided targeted repair.
By transforming long-form generation into a staged decision process, DramaAgent maintains reusable story and character states across scenes, improving controllability, character persistence, and cross-scene coherence while remaining compatible with diverse video backbones.
As illustrated in Fig.~\ref{fig:teaser}, DramaAgent generates long-form animated stories from a single narrative prompt by decomposing the story into controllable scene-level generation targets.

DramaAgent is built on three complementary components.
First, \textbf{hierarchical story planning} converts an open-ended narrative prompt into ordered scene-level targets, where each target specifies the participating characters, environment, event, emotional state, and transition cue.
Second, \textbf{persistent character conditioning} constructs reusable character references and carries them across scene generation, candidate selection, and repair, so that character identity is maintained as a persistent control signal.
Third, \textbf{reflection-guided targeted repair} scores generated clips along identity, semantic, temporal, and audio-visual dimensions, and regenerates low-quality scenes using constraints derived from the detected failure type.
These components allow DramaAgent to coordinate short-form generators under a long-form storytelling objective.
Instead of treating each clip as an independent generation result, DramaAgent maintains story context across scenes, selects candidates with respect to both local scene requirements and previous accepted content, and repairs local inconsistencies before composing the final audiovisual story.

We evaluate DramaAgent using both automatic and human-centered protocols.
Beyond standard visual quality metrics, we examine long-horizon coherence, character consistency, narrative fidelity, scene-level audio-visual consistency, and human preference.
Experiments across multiple video backbones show that DramaAgent consistently improves story-level coherence, character consistency, and narrative fidelity over direct generation and strong baselines.
Together, these results suggest that hierarchical agentic control provides a practical and model-agnostic way to improve controllable long-form audiovisual storytelling.
Our contributions are as follows:
\begin{itemize}
    \item We propose \textbf{DramaAgent}, a hierarchical and model-agnostic agentic framework for long-form text-to-video-and-audio generation over heterogeneous backbones.
    \item We introduce \textbf{persistent story-character state} and \textbf{reflection-guided targeted repair} to preserve cross-scene identity and narrative context while selectively regenerating problematic clips.
    \item Experiments across multiple video backbones demonstrate improved long-horizon coherence, character consistency, narrative fidelity, and scene-level audio-visual consistency over direct generation and strong baselines.
\end{itemize}

\begin{figure*}[htbp]
  \centering
  \includegraphics[width=1\linewidth]{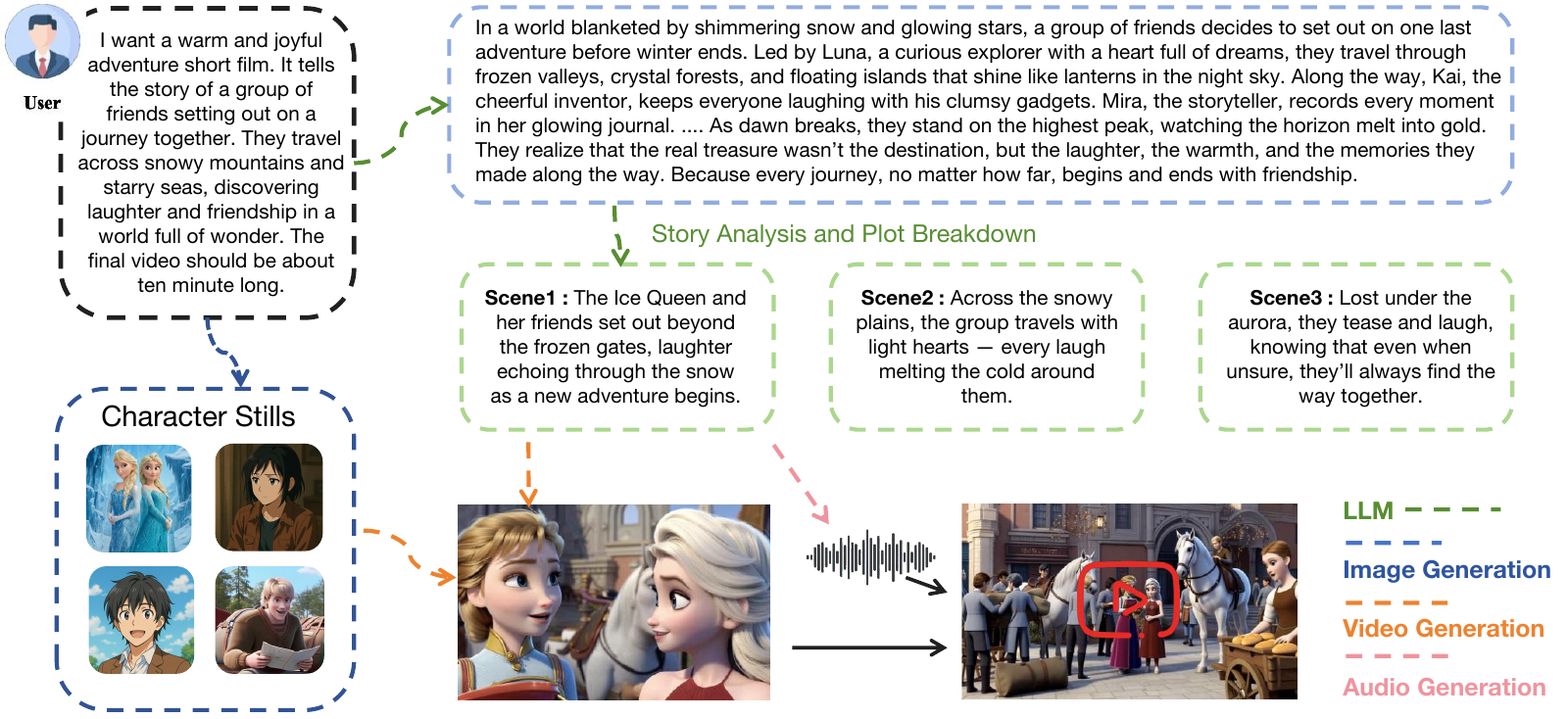}
  \caption{\textbf{Overview of DramaAgent.}
  DramaAgent decomposes a narrative prompt into scene-level targets, constructs persistent character stills, and coordinates image, video, and audio generators through scene-wise generation and reflection-guided repair.
  }
  \label{fig:example}
  \vspace{-0.6cm}
\end{figure*}

\section{Related Work}
\noindent\textbf{Long-form and agentic story video generation.}
Recent text-to-video research has moved beyond short clips toward story-level and long-form generation, where narrative coherence, character consistency, and temporal continuity across scenes become central challenges.
Methods such as StoryDiffusion~\cite{zhou2024storydiffusion}, AutoStory~\cite{wang2025autostory}, MovieFactory~\cite{zhu2023moviefactory}, and VideoStudio~\cite{long2024videostudio}, together with benchmarks such as MovieBench~\cite{wu2024moviebench}, highlight the need to preserve characters, events, and transitions over extended horizons.
Recent agentic frameworks such as StoryAgent~\cite{hu2024storyagent} and MovieAgent~\cite{wu2025movieagent} further improve controllability by decomposing narratives into scripts, storyboards, or scene-level prompts.
DramaAgent is related to these works, but focuses on long-form text-to-video-and-audio generation as a hierarchical control problem over heterogeneous generators.
It maintains reusable story and character states across scene generation, candidate selection, and repair, and uses reflection-guided diagnosis to perform failure-specific regeneration along identity, semantic, temporal, and audio-visual dimensions.

\noindent\textbf{Character-consistent video generation.}
Maintaining stable character identity is a longstanding challenge in long-form generation.
Recent methods address this problem through reference-image conditioning, subject-driven generation, personalized visual prompting, or keyframe-based control.
For example, Magic-Me~\cite{ma2024magic} shows that external visual references can improve subject consistency in generated videos.
Related story-generation systems also use storyboard images, keyframes, or visual anchors to stabilize character appearance across scenes.
DramaAgent builds on these efforts by treating character references as persistent control signals, propagating fixed references across scene-wise generation, candidate selection, and reflection-guided repair to reduce identity drift over long horizons.

\noindent\textbf{General video generation backbones.}  
Recent video generation methods have advanced rapidly with diffusion-based, autoregressive, and hybrid architectures.
Diffusion-based models such as VDM~\cite{videodiffusionmodel}, SEINE~\cite{chen2023seine}, I2VGen-XL~\cite{zhang2023i2vgen}, SVD~\cite{svd}, LaVie~\cite{wang2024lavie}, and MotionDirector~\cite{zhao2024motiondirector} improve visual realism, motion quality, and text alignment.
Large-scale and sequential generation models, including Sora~\cite{videoworldsimulators2024}, HunyuanVideo~\cite{kong2024hunyuanvideo}, VideoGPT~\cite{yan2021videogpt}, VideoPoet~\cite{kondratyuk2023videopoet}, and CogVideoX~\cite{yang2024cogvideox}, further strengthen video synthesis and temporal modeling.
While these models provide powerful short-form generation backbones, long-form storytelling still requires cross-scene state tracking, character persistence, narrative continuity, and audio-visual coordination.
DramaAgent is complementary to this line of work by coordinating heterogeneous video and audio generators under a long-form storytelling objective.

\section{Method}
We propose \textbf{DramaAgent}, a hierarchical and model-agnostic framework for long-form text-to-video-and-audio generation.
As shown in Fig.~\ref{fig:example}, DramaAgent coordinates heterogeneous video and audio generators through story decomposition, persistent character references, scene-wise generation, and reflection-guided selection and repair.

\subsection{Task Formulation}
Given a narrative prompt $P$, our goal is to generate a long-form video $\mathcal{V}$ with audio tracks $\mathcal{A}$ that preserve story coherence, character consistency, cross-scene continuity, and scene-level audio-visual consistency.
DramaAgent assumes access to a set of heterogeneous generation modules
$
    \mathcal{G}=\{g_1,g_2,\dots,g_M\},
$
where each module may correspond to a commercial API or an open-source backbone.
From $P$, DramaAgent constructs a temporally ordered scene sequence
$
    \mathcal{S}=\{S_1,S_2,\dots,S_N\},
$
and a persistent character reference set
$
    \mathcal{R}=\{R_c \mid c \in \mathcal{C}\},
$
where $\mathcal{C}$ denotes the set of story characters.
For each scene $S_i$, the framework generates candidate clips, evaluates them with respect to the scene specification and previous accepted context, and selectively repairs low-quality outputs.
The accepted video and audio outputs for scene $S_i$ are denoted by $v_i$ and $a_i$, respectively.
The final output is obtained by composing all accepted scene-level audiovisual segments:
$
    (\mathcal{V},\mathcal{A})=\mathrm{Compose}(\{(v_i,a_i)\}_{i=1}^{N}).
$

\subsection{Story Analysis and Plot Breakdown}
To support controllable long-form generation, DramaAgent first converts the input prompt into a structured story representation.
Given a narrative prompt $P$, an instruction-tuned language model expands it into an explicit story draft
$
    \mathcal{N}=\{n_1,n_2,\dots,n_L\},
$
which specifies key narrative elements such as environments, characters, major events, and emotional progression.
DramaAgent then decomposes $\mathcal{N}$ into a temporally ordered scene sequence $\mathcal{S}$.
Each scene corresponds to a coherent event, transition, or emotional beat.
Each scene is represented as
\begin{equation}
     S_i = (x_i,\mathcal{C}_i,e_i,m_i,u_i),
\end{equation}
where $x_i$ denotes the scene description, $\mathcal{C}_i$ the participating characters, $e_i$ the environment, $m_i$ the narrative role or emotional state, and $u_i$ an optional transition cue from the preceding scene.
These structured scene targets provide the semantic backbone for both video and audio generation.
Representative prompt templates are provided in \textit{App.}~\ref{app:prompt_reflection}.

\subsection{Persistent Character Conditioning}
After plot breakdown, DramaAgent constructs persistent character references, referred to as \textbf{Character Stills}, to stabilize character identity across scenes.
From the expanded story, the framework extracts the character set $\mathcal{C}$ and their identity-defining attributes, including appearance, clothing, hairstyle, age, and style cues.
For each character $c\in\mathcal{C}$, DramaAgent forms a character specification $A_c$ and generates a canonical visual reference:
\begin{equation}
    R_c=\mathrm{T2I}(A_c),
\end{equation}
where $\mathrm{T2I}(\cdot)$ denotes a text-to-image generator.

These references form a fixed character sheet and are reused throughout the pipeline.
For each scene $S_i$, the scene condition is augmented as
\begin{equation}
    \tilde{S}_i=(S_i,\{R_c \mid c\in\mathcal{C}_i\}),
\end{equation}
where $\mathcal{C}_i\subseteq\mathcal{C}$ denotes the characters appearing in scene $i$.
When a video generator supports image conditioning, the corresponding character stills are provided as visual inputs.
Otherwise, DramaAgent converts them into textual identity constraints.
By propagating the same references across generation, candidate selection, and repair, DramaAgent treats character identity as a persistent control signal rather than a one-time prompt attribute.

\subsection{Reflection-guided Candidate Selection and Repair}
Given the structured scene condition $\tilde{S}_i$, DramaAgent performs scene-wise generation over heterogeneous backbones.
For each scene, the framework produces a candidate set
\begin{equation}
    \mathcal{V}_i=\{v_i^{(1)},v_i^{(2)},\dots,v_i^{(K)}\},
\end{equation}
where candidates may come from different backbones, prompt variants, or random seeds under a fixed generation budget.
The accepted history before scene $i$ is summarized as
\begin{equation}
    h_{i-1}=\mathrm{Summarize}(\{S_j,v_j,a_j\}_{j<i}),
\end{equation}
which provides cross-scene context for candidate evaluation and repair.

\noindent\textbf{Candidate scoring.}
Each candidate is evaluated along four dimensions: identity consistency, semantic fidelity, temporal continuity, and audio-visual compatibility.
Identity consistency $s_{\mathrm{id}}$ measures whether visible characters match the persistent character references.
Semantic fidelity $s_{\mathrm{sem}}$ measures whether the required entities, actions, environments, and events in $\tilde{S}_i$ are present.
Temporal continuity $s_{\mathrm{temp}}$ measures whether the candidate is consistent with the accepted history $h_{i-1}$, including scene transition and story progression.
Audio-visual compatibility $s_{\mathrm{av}}$ measures whether the generated visual content is compatible with the dialogue, sound events, and audio plan associated with the scene.

The final reflection score is computed as
\begin{equation}
\label{equscore}
\begin{aligned}
\mathrm{Score}(v_i^{(k)}) =\;&
\lambda_{\mathrm{id}}\, s_{\mathrm{id}}(v_i^{(k)},\mathcal{R}) \\
&+ \lambda_{\mathrm{sem}}\, s_{\mathrm{sem}}(v_i^{(k)},\tilde{S}_i) \\
&+ \lambda_{\mathrm{temp}}\, s_{\mathrm{temp}}(v_i^{(k)},h_{i-1}) \\
&+ \lambda_{\mathrm{av}}\, s_{\mathrm{av}}(v_i^{(k)},D_i),
\end{aligned}
\end{equation}
where $D_i$ denotes the scene-level dialogue and audio plan.
In the default setting, we use fixed weights
$\lambda_{\mathrm{id}}=0.35$,
$\lambda_{\mathrm{sem}}=0.30$,
$\lambda_{\mathrm{temp}}=0.20$, and
$\lambda_{\mathrm{av}}=0.15$.
The detailed scoring prompts and rubric are provided in \textit{App.}~\ref{app:reflection_settings}.

\noindent\textbf{Targeted repair.}
The best candidate is selected by
\begin{equation}
    v_i^\star=\arg\max_{v_i^{(k)}\in\mathcal{V}_i}\mathrm{Score}(v_i^{(k)}).
\end{equation}
If $\mathrm{Score}(v_i^\star)\ge\tau$, the scene is accepted.
Otherwise, DramaAgent identifies the weakest evaluation dimension:
\begin{equation}
    d_i^\star=\arg\min_{d\in\{\mathrm{id},\mathrm{sem},\mathrm{temp},\mathrm{av}\}} s_d(v_i^\star),
\end{equation}
and rewrites the scene prompt according to the corresponding failure type.

For identity failures, the repair prompt strengthens character-reference constraints.
For semantic failures, it injects missing entities, actions, or environmental details.
For temporal failures, it adds transition cues from the accepted history.
For audio-visual failures, it adjusts dialogue timing, sound-event constraints, or visual pacing.
The repaired scene is then regenerated and rescored.

This process is repeated for at most $R$ rounds or until the score improvement falls below $\epsilon$:
\begin{equation}
    \mathrm{Score}(v_{i,t}^\star)-\mathrm{Score}(v_{i,t-1}^\star)<\epsilon.
\end{equation}
We use $\tau=0.80$, $\epsilon=0.02$, and $R=2$ by default.
This mechanism turns refinement into failure-aware correction rather than generic resampling, allowing DramaAgent to repair local inconsistencies before they propagate to later scenes.

\subsection{Audio Generation and Composition}
To complement visual generation, DramaAgent performs scene-aware audio generation for dialogue, ambient sound, and background music.
Given the scene sequence $\mathcal{S}$ and character set $\mathcal{C}$, a language model first produces dialogue scripts
$
    \mathcal{D}=\{D_1,D_2,\dots,D_N\},
$
where each $D_i$ is conditioned on the current scene content, previous story context, participating characters, emotional state, and estimated clip duration.
Each dialogue line is annotated with speaker identity, emotion, and delivery style for downstream voice assignment and prosody control.

DramaAgent then uses a multi-speaker text-to-speech system to synthesize character-specific speech tracks.
Ambient sound and background music are generated or retrieved according to the scene environment and narrative role.
Finally, the accepted scene clips $\{v_i\}$ and audio tracks $\{a_i\}$ are assembled into the final long-form audiovisual output.
Scene boundaries are smoothed with visual and auditory transitions, and dialogue timing is aligned with clip duration as closely as possible.

The audio module mainly supports scene-level audio-visual consistency by providing dialogue, sound-event, and pacing constraints for generation and reflection.
Fine-grained lip synchronization and precise sound-effect timing still depend on the underlying video and audio generators, and are discussed as limitations.

\begin{table*}[ht]
  \centering
  \caption{\textbf{Main comparison with recent text-to-video generation methods.}
  Results are reported for script synopsis to keyframe/storyboard generation and script synopsis to full movie generation.
  DramaAgent achieves strong overall performance across both settings.
  }
  \resizebox{\linewidth}{!}{
  \begin{tabular}{l|cc|ccccc}
    \toprule
    \multirow{2}{*}{Method} & \multirow{2}{*}{CLIP$\uparrow$} & \multirow{2}{*}{Inception$\uparrow$} & \multicolumn{5}{c}{VBench Metrics /\% $\uparrow$} \\
    \cline{4-8}
    & & & Subject Cons. & Bg Cons. & Motion Smth. & Dyn. Degree & Aesthetic \\
    \midrule
    \multicolumn{8}{l}{\textit{Script Synopsis to Keyframe/StoryBoard Generation}} \\
    \midrule
    StoryDiffusion~\cite{zhou2024storydiffusion} & 20.46 & 6.24 & - & - & - & - & - \\
    AutoStory~\cite{wang2025autostory} & 20.21 & 6.01 & - & - & - & - & - \\
    MovieAgent~\cite{wu2025movieagent} & 22.12 & 7.23 & - & - & - & - & - \\ \midrule
    \textbf{DramaAgent (Ours)}  & \textbf{23.02} & \textbf{7.64} & - & - & - & - & - \\
    \midrule
    \multicolumn{8}{l}{\textit{Script Synopsis to Movie Generation}} \\
    \midrule
    StoryDiffusion~\cite{zhou2024storydiffusion} + SVD~\cite{svd} & 21.39 & 8.36 & 93.64 & 93.78 & 96.30 & 74.48 & 56.69 \\
    StoryDiffusion~\cite{zhou2024storydiffusion}+ CogVideoX~\cite{yang2024cogvideox} & 21.83 & 9.01 & 93.45 & 94.56 & 96.60 & 27.89 & 56.05 \\
    AutoStory~\cite{wang2025autostory} + CogVideoX~\cite{yang2024cogvideox} & 20.27 & 7.21 & 91.45 & 93.32 & 95.87 & 70.32 & 52.34 \\
    DreamVideo~\cite{wei2024dreamvideo} & 21.37 & 8.11 & 93.17 & 93.77 & 96.40 & 26.97 & 42.16 \\
    Magic-Me~\cite{ma2024magic} & 21.72 & 8.34 & 94.01 & 94.68 & 96.41 & 14.86 & 55.89 \\
    MovieAgent~\cite{wu2025movieagent} & 22.25 & 9.39 & 94.72 & 93.52 & 97.84 & \textbf{76.27} & 58.63 \\
    \midrule
    \textbf{DramaAgent (Ours)}  & \textbf{24.34} & \textbf{9.67} & \textbf{97.02} & \textbf{97.54} & \textbf{98.14} & 67.32 & \textbf{64.91} \\
    \bottomrule
  \end{tabular}
  }
  \label{tab:method_comparison}
  \vspace{-0.6cm}
\end{table*}

\section{Experiments}
We evaluate DramaAgent from four perspectives: overall generation quality, robustness across heterogeneous video backbones, the contribution of each control component, and human-perceived long-form coherence.
Beyond visual quality, our evaluation focuses on whether the generated story preserves character identity, narrative progression, cross-scene continuity, and scene-level audio-visual consistency.

\subsection{Experimental Setup}
\label{sec:experiment_setup}
\noindent\textbf{Evaluation data.}
We evaluate DramaAgent on narrative-to-video and script-to-storyboard generation tasks.
The evaluation set contains 50 narrative scripts spanning multiple genres, including fantasy stories, conversational scenes, emotional monologues, and action sequences.
Each script contains 10--20 sentences and corresponds to approximately 30--60 seconds of video content after scene decomposition and composition.

\noindent\textbf{Generation backbones.}
For the video generation stage, we use heterogeneous backbones including Wan2.2-I2V-Flash, Keling, Jimeng, Hailuo~3.0, and Seedance.
These backbones are accessed through their corresponding generation interfaces, and DramaAgent adapts scene prompts and conditioning inputs to each interface.
To control the local generation setting, all methods use the same scene-level candidate count, clip duration, and resolution, i.e., 3 candidates per scene, 5-second clips, and 1280$\times$720 resolution.
Unless otherwise stated, DramaAgent performs at most 2 rounds of reflection-guided repair.
Detailed generation settings, backbone versions, API access dates, and reflection hyperparameters are provided in \textit{App.}~\ref{app:generation_details} and \textit{App.}~\ref{app:reflection_settings}.

\noindent\textbf{Metrics.}
Following the VBench protocol~\cite{huang2024vbench}, we report \textit{Subject Consistency}, \textit{Background Consistency}, \textit{Motion Smoothness}, \textit{Dynamic Degree}, and \textit{Aesthetic} quality.
We additionally report \textit{CLIP} and \textit{Inception} scores to assess semantic alignment and visual realism.
Since automatic visual metrics cannot fully capture long-form storytelling quality, we also conduct blind human evaluation on temporal coherence, character consistency, narrative fidelity, and scene-level audio-visual consistency.
The human-evaluation protocol and participant statistics are provided in \textit{App.}~\ref{app:human_eval}.
Runtime and approximate API cost statistics are reported in \textit{App.}~\ref{app:runtime_cost} and Tab.~\ref{tab:runtime_cost}.

\begin{figure*}[h]
  \centering
  \includegraphics[width=\linewidth]{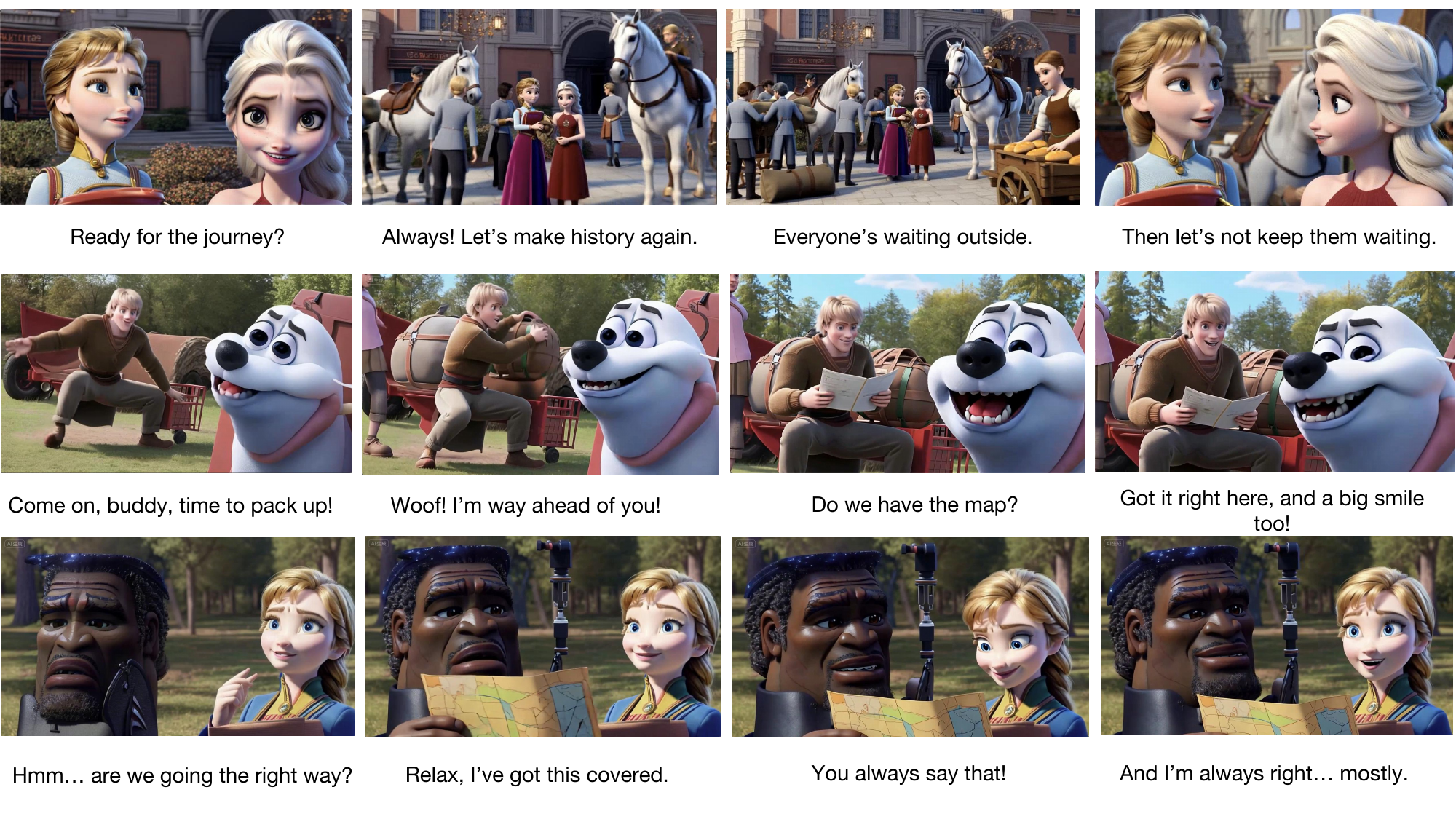}
  \caption{\textbf{Qualitative results across three narrative scenes.}
  DramaAgent maintains recurring characters and coherent scene progression across different story contexts.
  The examples illustrate the role of structured planning, persistent conditioning, and reflection-guided repair in long-form storytelling.
  }
  \label{fig:qualitative_results}
  \vspace{-0.6cm}
\end{figure*}

\subsection{Main Results}
Tab.~\ref{tab:method_comparison} compares DramaAgent with recent text-to-video and story video generation methods under two settings: \textit{Script Synopsis to Keyframe/Storyboard Generation} and \textit{Script Synopsis to Movie Generation}.
DramaAgent achieves strong overall performance in both settings.

\noindent\textbf{Script synopsis to keyframe/storyboard generation.}
In the storyboard generation setting, DramaAgent obtains a CLIP score of 23.02 and an Inception score of 7.64, outperforming StoryDiffusion, AutoStory, and MovieAgent.
These results indicate that structured story planning and character-aware generation produce more semantically grounded intermediate visual targets from high-level narrative synopses.
Since storyboard quality directly affects downstream video generation, this setting evaluates whether DramaAgent can provide a stronger visual foundation for long-form synthesis.

\begin{table}[t]
\centering
\caption{\textbf{Component ablation on a fixed backbone.}
Progressively adding story planning, character stills, and reflective repair on Seedance.}
\resizebox{\linewidth}{!}{
\begin{tabular}{l|cc|ccccc}
\toprule
\multirow{2}{*}{Method} 
& \multirow{2}{*}{CLIP$\uparrow$} 
& \multirow{2}{*}{Inception$\uparrow$} 
& \multicolumn{5}{c}{VBench Metrics / \% $\uparrow$} \\
\cline{4-8}
& & & Subject Cons. & Bg Cons. & Motion Smth. & Dyn. Degree & Aesthetic \\
\midrule
Direct Generation (Seedance) & 22.90 & 8.65 & 92.10 & 95.60 & 97.10 & 64.20 & 62.30 \\
+ Story Planning              & 23.55 & 8.95 & 93.80 & 96.30 & 97.45 & 65.10 & 62.90 \\
+ Character Stills            & 24.12 & 9.41 & 96.85 & 97.33 & 97.96 & 66.84 & 64.37 \\
+ Reflective Repair (Full)    & 24.34 & 9.67 & 97.02 & 97.54 & 98.14 & 67.32 & 64.91 \\
\bottomrule
\end{tabular}
}
\label{tab:fixed_backbone_ablation}
\vspace{-0.6cm}
\end{table}
\noindent\textbf{Script synopsis to movie generation.}
In the full movie generation setting, DramaAgent reaches a CLIP score of 24.34 and an Inception score of 9.67.
It also obtains the highest scores in Subject Consistency, Background Consistency, Motion Smoothness, and Aesthetic quality among the compared methods.
Compared with StoryDiffusion+SVD, StoryDiffusion+CogVideoX, AutoStory+CogVideoX, DreamVideo, Magic-Me, and MovieAgent, DramaAgent produces stronger semantic alignment and more stable character consistency.
These results suggest that structured control is effective for organizing scene-level generations into a coherent long-form story.

\subsection{Controlled Analysis}

To isolate the contribution of DramaAgent from backbone choice and additional generation attempts, we conduct fixed-backbone and generation-budget-aware analyses.

\noindent\textbf{Fixed-backbone analysis.}
Tab.~\ref{tab:fixed_backbone_ablation} reports a progressive ablation using the same video backbone.
Adding story planning, character stills, and reflection-guided repair progressively improves performance, showing that the gains arise from the proposed control components rather than the backbone alone.

\noindent\textbf{Generation-budget-aware comparison.}
Since candidate selection and repair may introduce additional generation calls, we report the local generation setting, runtime/cost statistics, and controlled variants with and without reflection.
This helps clarify the quality-cost tradeoff and separates targeted repair from generic additional sampling.

\begin{figure*}[h]
  \centering
  \includegraphics[width=0.95\linewidth]{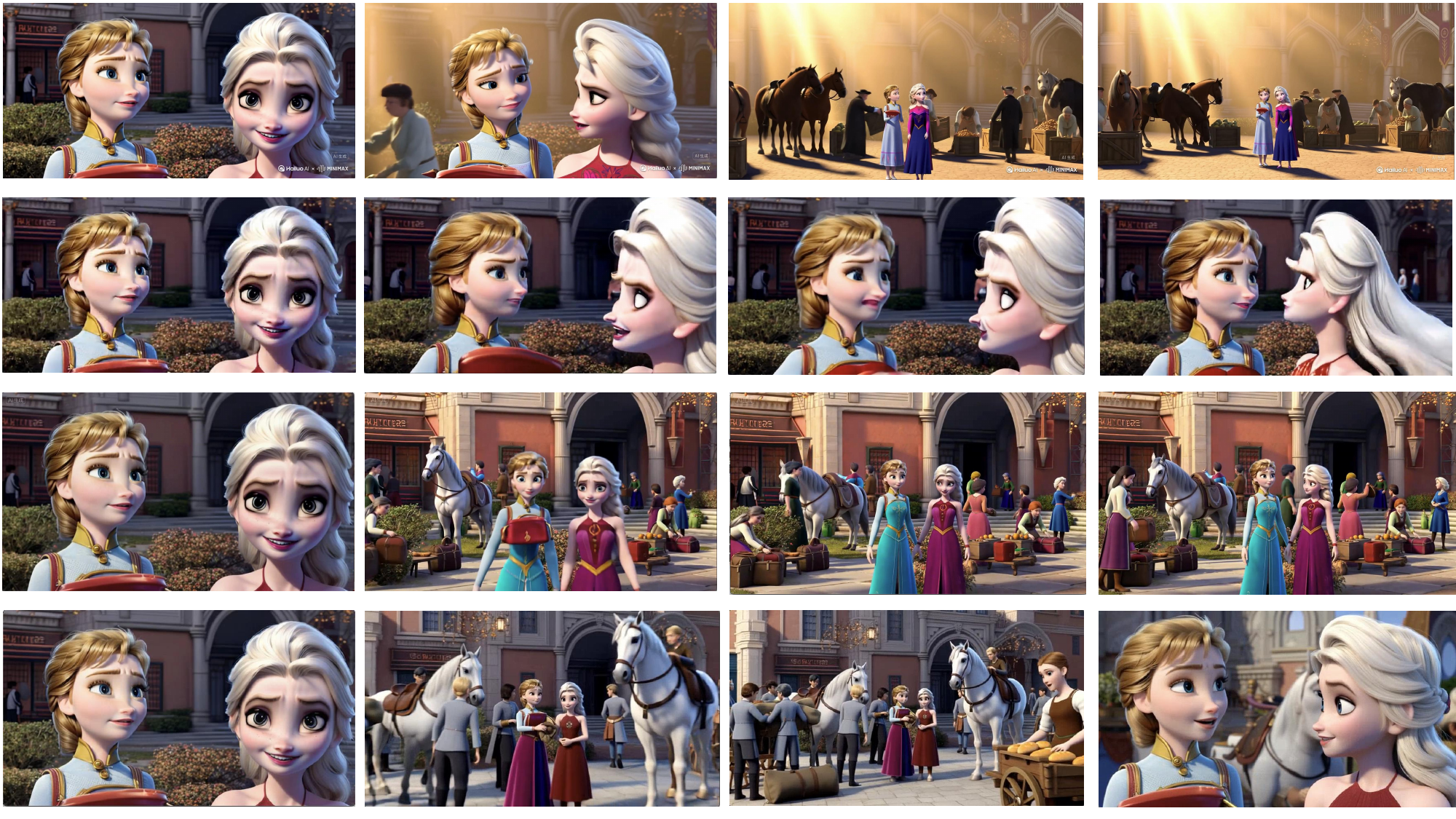}
  \caption{\textbf{Qualitative comparison across four video generation backbones.}
  We compare Hailuo, Keling, Jimeng, and SeedDance under the same narrative prompt and character references, highlighting their differences in identity consistency, temporal stability, and scene realism.}
  \label{fig:api_comparison_v2}
  \vspace{-0.6cm}
\end{figure*}

\subsection{Ablation Study}
\noindent\textbf{Effect of backbone choice.}
Tab.~\ref{tab:method_comparison1} compares DramaAgent with different video generation backbones.
Although performance varies across generators, DramaAgent remains consistently competitive, suggesting that the framework is not tied to a specific backbone.
Seedance achieves the best balance between semantic alignment and aesthetic quality, while Jimeng and Hailuo~3.0 perform strongly on motion-related metrics.

\begin{table}[t]
  \centering
  \caption{\textbf{Effect of backbone choice.}
  DramaAgent is evaluated with five video generation backbones under the same control framework.}
  \resizebox{\linewidth}{!}{
  \begin{tabular}{l|cc|ccccc}
    \toprule
    \multirow{2}{*}{Method} 
    & \multirow{2}{*}{CLIP$\uparrow$} 
    & \multirow{2}{*}{Inception$\uparrow$} 
    & \multicolumn{5}{c}{VBench Metrics / \% $\uparrow$} \\
    \cline{4-8}
    & & & Subject Cons. & Bg Cons. & Motion Smth. & Dyn. Degree & Aesthetic \\
    \midrule
    DramaAgent (Wan2.2-I2V-Flash) & 23.47 & \textbf{9.56} & 94.12 & 93.95 & 96.92 & 70.15 & 54.32 \\
    DramaAgent (Keling)            & 23.78 & 9.35 & 95.64 & 95.82 & 97.43 & 66.92 & 58.14 \\
    DramaAgent (Jimeng)            & 23.96 & 9.38 & 96.42 & 96.71 & 98.02 & 80.45 & 61.27 \\
    DramaAgent (Hailuo~3.0)        & 22.91 & 9.12 & 96.34 & 97.27 & \textbf{98.61} & \textbf{84.31} & 61.32 \\
    DramaAgent (Seedance)          & \textbf{24.12} & 9.41 & \textbf{96.85} & \textbf{97.33} & 97.96 & 66.84 & \textbf{64.37} \\
    \bottomrule
  \end{tabular}
  }
  \label{tab:method_comparison1}
  \vspace{-0.6cm}
\end{table}

\noindent\textbf{Effect of reflective refinement.}
Tab.~\ref{tab:method_comparison2} shows that reflection consistently improves DramaAgent in both storyboard and full movie generation.
It raises CLIP from 22.05 to 23.02 for storyboard generation, and improves CLIP/Inception from 24.12/9.41 to 24.34/9.67 for full movie generation, with gains across VBench metrics.
These results show that reflection-guided repair strengthens both intermediate grounding and final long-form synthesis.

\begin{table}[t]
  \centering
  \caption{\textbf{Effect of reflective refinement.}
  Comparison with and without reflection in storyboard and full movie generation.}
  \resizebox{\linewidth}{!}{
  \begin{tabular}{l|cc|ccccc}
    \toprule
    \multirow{2}{*}{Method} 
    & \multirow{2}{*}{CLIP$\uparrow$} 
    & \multirow{2}{*}{Inception$\uparrow$} 
    & \multicolumn{5}{c}{VBench Metrics / \% $\uparrow$} \\
    \cline{4-8}
    & & & Subject Cons. & Bg Cons. & Motion Smth. & Dyn. Degree & Aesthetic \\
    \midrule
    \multicolumn{8}{l}{\textit{Script Synopsis to Keyframe/Storyboard Generation}} \\
    DramaAgent w/o Reflection & 22.05 & \textbf{7.64} & - & - & - & - & - \\
    DramaAgent w/ Reflection  & \textbf{23.02} & \textbf{7.64} & - & - & - & - & - \\
    \midrule
    \multicolumn{8}{l}{\textit{Script Synopsis to Movie Generation}} \\
    DramaAgent w/o Reflection & 24.12 & 9.41 & 96.85 & 97.33 & 97.96 & 66.84 & 64.37 \\
    DramaAgent w/ Reflection  & \textbf{24.34} & \textbf{9.67} & \textbf{97.02} & \textbf{97.54} & \textbf{98.14} & \textbf{67.32} & \textbf{64.91} \\
    \bottomrule
  \end{tabular}
  }
  \label{tab:method_comparison2}
  \vspace{-0.6cm}
\end{table}

\noindent\textbf{Component ablation on a fixed backbone.}
Tab.~\ref{tab:fixed_backbone_ablation} reports a progressive ablation on the same backbone.
Story planning improves semantic alignment, character stills strengthen identity preservation, and reflective repair adds further gains, validating the complementary roles of the three components.

\subsection{Audio-Visual and Human Evaluation}
Automatic visual metrics alone cannot fully capture long-form audiovisual storytelling quality.
We therefore conduct a blind human evaluation on temporal coherence, character consistency, narrative fidelity, and scene-level audio-visual consistency.
Human results show that DramaAgent is preferred over baseline pipelines across these dimensions, indicating stronger scene continuity, character persistence, and semantic compatibility between audio and visual content.
Detailed results are reported in Tab.~\ref{tab:human_eval} in \textit{App.}~\ref{app:human_eval}.
This evaluation focuses on scene-level audio-visual consistency rather than perfect frame-level lip synchronization, which remains partly constrained by the underlying generators.

\subsection{Qualitative Analysis}
\noindent\textbf{Cross-scene storytelling.}
Fig.~\ref{fig:qualitative_results} shows generated results across three narrative scenes.
DramaAgent maintains recurring character appearances and coherent scene progression across different story contexts.
These examples illustrate the role of structured planning, persistent character conditioning, and reflection-guided repair in long-form storytelling.
Additional case studies are provided in \textit{App.}~\ref{app:qualitative}.

\noindent\textbf{Cross-backbone comparison.}
Fig.~\ref{fig:api_comparison_v2} compares different video generation backbones under the same narrative prompt and character references.
The results show complementary strengths across backbones, supporting the model-agnostic design of DramaAgent.
For example, some backbones produce stronger motion continuity, while others better preserve character appearance or visual style.

\section{Conclusion}
We propose \textbf{DramaAgent}, a hierarchical and model-agnostic framework for long-form text-to-video-and-audio generation.
DramaAgent formulates long-form storytelling as a structured control problem over heterogeneous video and audio generators, combining story planning, persistent character conditioning, scene-wise generation, reflection-guided repair, and audio composition.
Experiments across multiple backbones show that DramaAgent improves long-horizon coherence, character consistency, narrative fidelity, and scene-level audio-visual consistency over direct generation and strong baselines.
These results suggest that hierarchical agentic control is a practical direction for controllable long-form audiovisual storytelling.

\section*{Limitations}
DramaAgent has several limitations.
First, it relies on third-party video and audio generation APIs, so outputs, latency, pricing, and conditioning interfaces may change over time.
Second, as an orchestration framework rather than a unified world-consistent generative model, it may still struggle with complex interactions, physical contact, large pose/camera changes, and residual identity drift.
Third, DramaAgent improves scene-level audio-visual consistency but does not fully solve frame-level lip-sync or precise sound-effect timing, which depend on the underlying generators.
Finally, generated videos may inherit artifacts, biases, and safety risks from the underlying models, and should be clearly disclosed as synthetic content in real-world use.

\bibliography{customs}

\clearpage

\appendix

\section{Detailed Generation Settings}
\label{app:generation_details}

\begin{table*}[t]
\centering
\caption{\textbf{Reproducibility settings for video generation backbones.}
We report the backbone/API variant, access date, supported conditioning interface, and scene-level generation budget used in our experiments.
All DramaAgent variants share the same upper-level planning, persistent character conditioning, and reflective repair framework; only the underlying video generator changes across rows.}
\resizebox{\textwidth}{!}{
\begin{tabular}{l|l|l|l|c|c|c|c}
\toprule
Backbone & Version / API Variant & Access Date & Input Type & Candidates / Scene & Repair Rounds & Clip Duration & Resolution \\
\midrule
Wan2.2-I2V-Flash & Wan2.2-I2V-Flash & 2025-11 & image + text & 3 & 2 & 5s & 1280$\times$720 \\
Keling & Keling & 2025-11 & image + text & 3 & 2 & 5s & 1280$\times$720 \\
Jimeng & Jimeng & 2025-11 & text-only / multimodal & 3 & 2 & 5s & 1280$\times$720 \\
Hailuo 3.0 & Hailuo 3.0 & 2025-11 & image + text & 3 & 2 & 5s & 1280$\times$720 \\
Seeddance 3.0 & Seeddance 3.0 & 2025-11 & image + text & 3 & 2 & 5s & 1280$\times$720 \\
\bottomrule
\end{tabular}
}
\label{tab:repro_settings}
\end{table*}

Table~\ref{tab:repro_settings} summarizes the generation settings used in our experiments for reproducibility.
Unless otherwise stated, all methods are evaluated on the same set of 50 narrative scripts, where each script contains 10--20 sentences and corresponds to approximately 30--60 seconds of video content after scene decomposition and composition.

Across different backbones, the upper-level DramaAgent framework is kept unchanged, including structured story planning, persistent character conditioning, and reflective repair.
Only the underlying video generator is varied.
When a backbone supports direct image conditioning, character stills are provided as visual inputs; otherwise, the same references are converted into textual identity constraints and injected into the scene prompt.

\noindent\textbf{Reproducibility note.}
Because some commercial APIs may evolve over time, exact outputs can vary across access dates.
To improve reproducibility, we record the API variant and access date for each experiment, while keeping the scene-level budget, resolution, repair rounds, and upper-level DramaAgent logic fixed within each comparison group.

\subsection{Runtime and Cost Statistics}
\label{app:runtime_cost}

\begin{table}[t]
\centering
\caption{\textbf{Runtime and cost statistics under the default setting.}
Reported values are averages over the evaluation set and may vary with backbone API latency and access date.}
\resizebox{0.9\linewidth}{!}{
\begin{tabular}{l c}
\toprule
Statistic & Value \\
\midrule
Average initial generation time / scene & 75 s \\
Average repaired generation time / scene & 28 s \\
Average end-to-end time / 30--60s script & 18--32 min \\
Average API calls / scene & 4.1 \\
Approximate cost / minute of final video & \$7.8 \\
\bottomrule
\end{tabular}
}
\label{tab:runtime_cost}
\end{table}

Table~\ref{tab:runtime_cost} reports practical runtime and approximate API usage statistics for the default setting.
Under the default configuration ($K=3$ candidates per scene, at most $R=2$ repair rounds, 5-second clips, and 1280$\times$720 resolution), DramaAgent requires on average 75 seconds for initial scene generation and 28 seconds for each repaired scene.
Across the full pipeline, the average end-to-end generation time for a 30--60 second script is approximately 18--32 minutes, depending on the number of scenes, the activated repair rounds, and the selected backbone APIs.
On average, DramaAgent makes 4.1 API calls per scene, including candidate generation and optional regeneration calls.
The approximate generation cost is about \$7.8 per minute of final video under the default setup.
These values should be interpreted as practical reference numbers rather than fixed constants, since commercial API latency and pricing may vary across providers and access dates.

\section{Reflection Settings and Repair Policy}
\label{app:reflection_settings}
\begin{table}[t]
\centering
\caption{\textbf{Default reflection settings.}
We use the following default hyperparameters for candidate scoring and selective reflective repair.}
\resizebox{0.65\linewidth}{!}{
\begin{tabular}{l c}
\toprule
Parameter & Value \\
\midrule
$\lambda_{\mathrm{id}}$   & 0.35 \\
$\lambda_{\mathrm{sem}}$  & 0.30 \\
$\lambda_{\mathrm{temp}}$ & 0.20 \\
$\lambda_{\mathrm{av}}$   & 0.15 \\
Acceptance threshold $\tau$ & 0.80 \\
Early-stop threshold $\epsilon$ & 0.02 \\
Max repair rounds $R$ & 2 \\
Candidates per scene $K$ & 3 \\
\bottomrule
\end{tabular}
}
\label{tab:reflection_settings}
\end{table}
\noindent\textbf{Repair policy.}
Table~\ref{tab:reflection_settings} summarizes the default reflection hyperparameters.
For each scene, DramaAgent first generates $K=3$ candidate clips and computes the reflection score in Eq.~\ref{equscore}.
If the best candidate score is below $\tau=0.80$, the framework triggers targeted repair.
The repair route is chosen according to the weakest sub-score among
$s_{\mathrm{id}}$, $s_{\mathrm{sem}}$, $s_{\mathrm{temp}}$, and $s_{\mathrm{av}}$:
low identity consistency strengthens character-reference conditioning,
low semantic fidelity rewrites the prompt with missing entities or actions,
low temporal continuity injects stronger transition cues from the previous scene,
and low audio-visual compatibility adjusts dialogue timing or regenerates the clip under tighter temporal constraints.
We perform at most $R=2$ repair rounds per scene, and stop early when the score improvement is below $\epsilon=0.02$.

\section{Human Evaluation Protocol}
\label{app:human_eval}

This section describes the human-evaluation protocol used to complement automatic metrics for long-form story generation.

\noindent\textbf{Evaluation dimensions.}
We evaluate generated outputs on four human-centered dimensions: \textit{temporal coherence}, \textit{character consistency}, \textit{narrative fidelity}, and \textit{audio-visual alignment}. These dimensions are designed to capture aspects of long-form generation that are not fully reflected by standard automatic visual metrics.

\noindent\textbf{Participants.}
We recruit 15 human evaluators with prior experience in video content assessment, multimedia generation, or related visual-language tasks. All participants are blind to the method identities during evaluation.

\noindent\textbf{Evaluation protocol.}
We use blind pairwise comparison for generated samples. For each comparison, evaluators are shown outputs from two methods under the same narrative prompt and are asked to indicate the preferred result along the four evaluation dimensions. Each sample is evaluated independently by at least 3 annotators.

\noindent\textbf{Evaluation criteria.}
\textit{Temporal coherence} measures whether scene transitions, motion progression, and event continuity remain stable across time. \textit{Character consistency} measures whether the same characters preserve their visual identity across scenes. \textit{Narrative fidelity} measures whether the generated output remains faithful to the story content, event sequence, and emotional progression specified in the input prompt. \textit{Audio-visual alignment} measures whether dialogue, sound, and visual content remain perceptually synchronized and semantically consistent.

\begin{table}[t]
\centering
\caption{\textbf{Human evaluation results.} DramaAgent is preferred over baseline pipelines across all four evaluation dimensions.}
\resizebox{\linewidth}{!}{
\begin{tabular}{lc}
\toprule
\textbf{Metric} & \textbf{DramaAgent Preference} \\
\midrule
Temporal Coherence      & 66.7\% \\
Character Consistency   & 71.3\% \\
Narrative Fidelity      & 63.3\% \\
Audio-Visual Alignment  & 68.0\% \\
\midrule
Inter-rater Agreement (Fleiss' Kappa) & 0.42 \\
\bottomrule
\end{tabular}
}
\label{tab:human_eval}
\end{table}

\noindent\textbf{Statistics.}
For each evaluation dimension, we report the average preference rate of DramaAgent across all evaluated samples.
Inter-rater agreement is measured by Fleiss' kappa.
The detailed human-evaluation results, including preference rates for temporal coherence, character consistency, narrative fidelity, and audio-visual alignment, are reported in Table~\ref{tab:human_eval}.

\begin{figure*}[h]
  \centering
  \includegraphics[width=\linewidth]{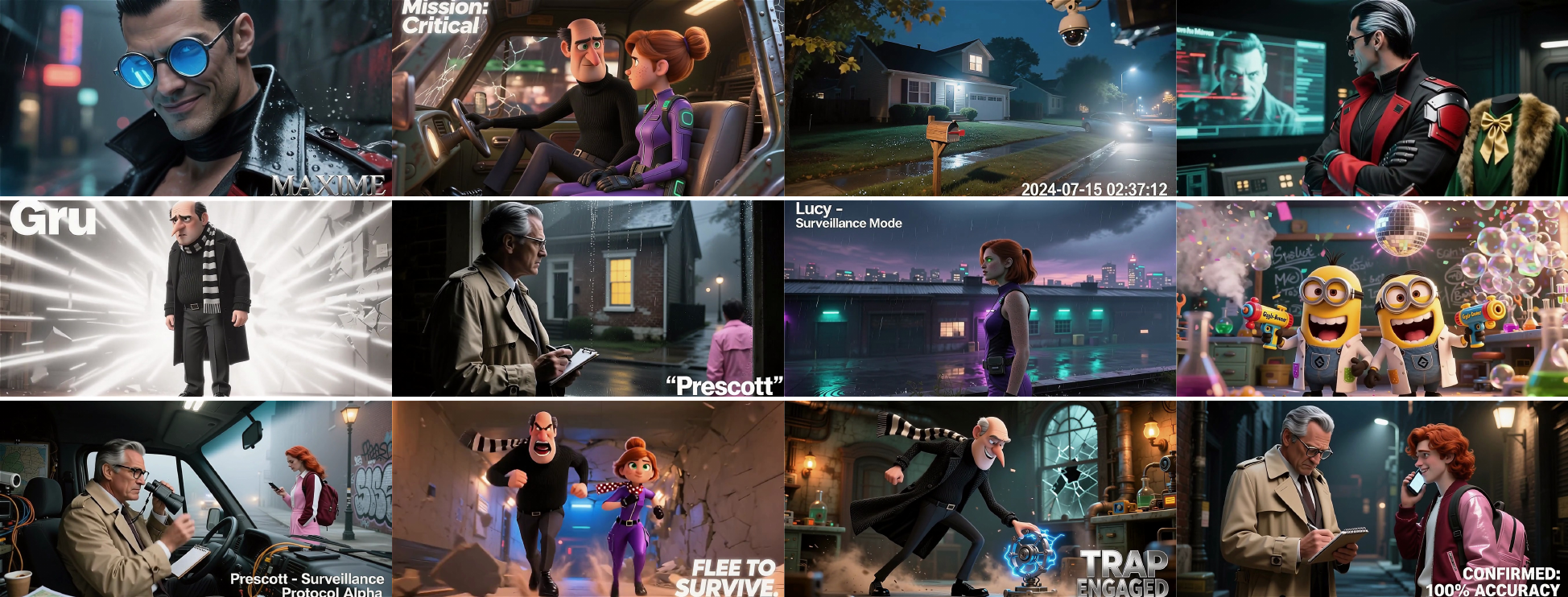}
  \caption{\textbf{Additional qualitative case: animated family-adventure narrative.}
  DramaAgent preserves recurring character identity, expressive interaction, and multi-scene story continuity in a stylized family-oriented animated story.}
  \label{fig:qual1}
\end{figure*}

\begin{figure*}[h]
  \centering
  \includegraphics[width=\linewidth]{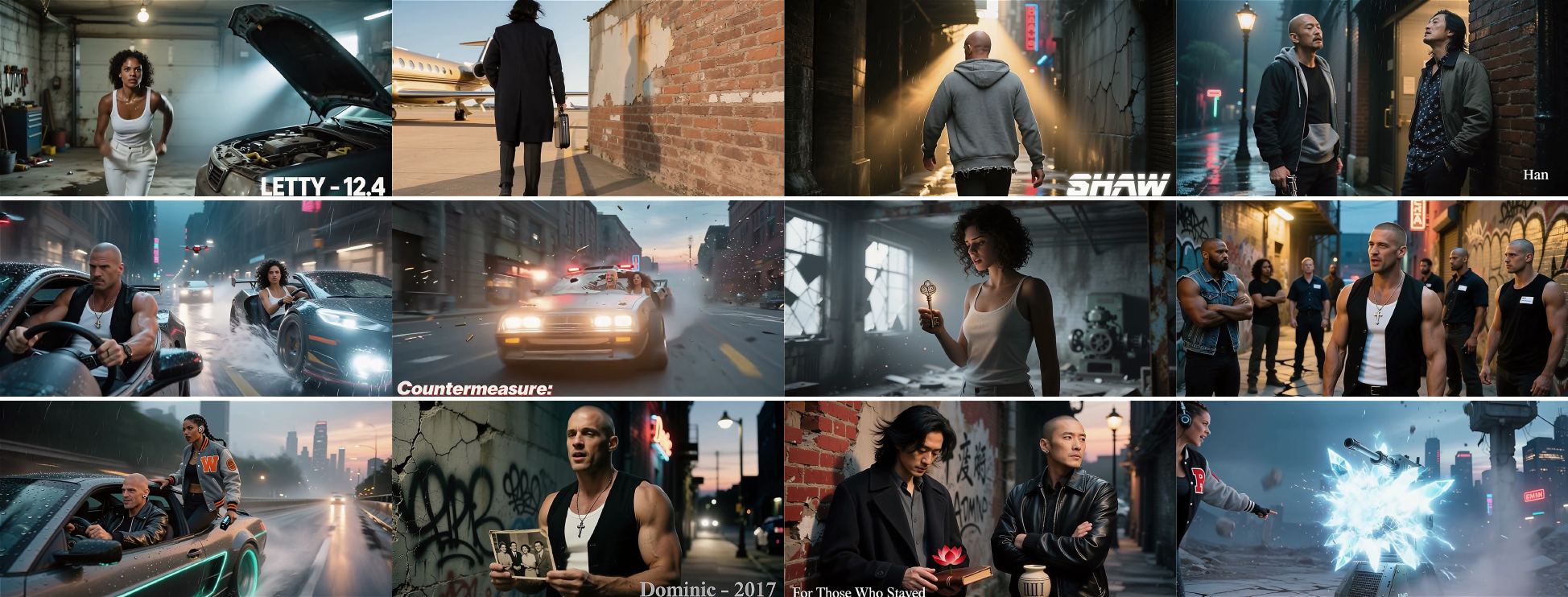}
  \caption{\textbf{Additional qualitative case: action-crime ensemble narrative.}
  The example illustrates multi-character coordination, dynamic scene progression, and stable identity preservation in a high-motion cinematic setting.}
  \label{fig:qual2}
\end{figure*}
\section{Prompt Templates and Reflection Settings}
\label{app:prompt_reflection}

This section provides representative prompt templates and key settings used in DramaAgent, including story analysis and plot breakdown, character still generation, reflective rewriting, and repair control parameters.

\subsection{Story Analysis and Plot Breakdown Prompt}

To convert a narrative prompt into a structured story representation, we use an instruction-tuned language model to expand the input into a longer story draft and decompose it into ordered scene-level units.
A representative prompt template is shown below.
\begin{quote}
\small
\textbf{Input:} A short narrative prompt describing the story theme, characters, and emotional tone.

\textbf{Instruction:} Expand the prompt into a coherent story with clear event progression. Then decompose the story into a sequence of scene-level descriptions. Each scene should correspond to a semantically coherent event or emotional beat. For each scene, provide: (1) a concise scene description, (2) participating characters, (3) environment, (4) emotional state or narrative role, and (5) a transition cue if needed.
\end{quote}

The output of this stage is used to construct the scene sequence $\mathcal{S}=\{S_1,S_2,\dots,S_N\}$ in the main pipeline.

\begin{figure*}[h]
  \centering
  \includegraphics[width=\linewidth]{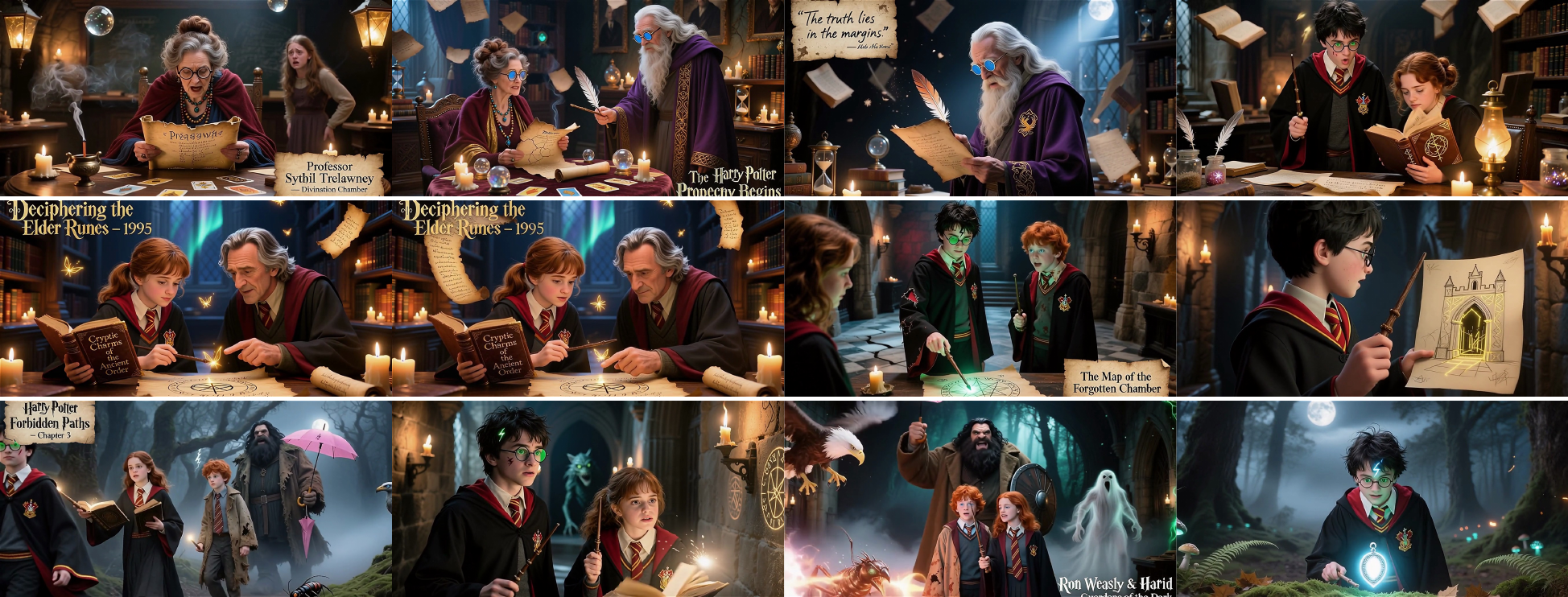}
  \caption{\textbf{Additional qualitative case: fantasy mystery narrative.}
  DramaAgent supports atmosphere-heavy story generation with coherent clue progression, recurring character consistency, and visually stable magical world-building.}
  \label{fig:qual3}
\end{figure*}

\begin{figure*}[h]
  \centering
  \includegraphics[width=\linewidth]{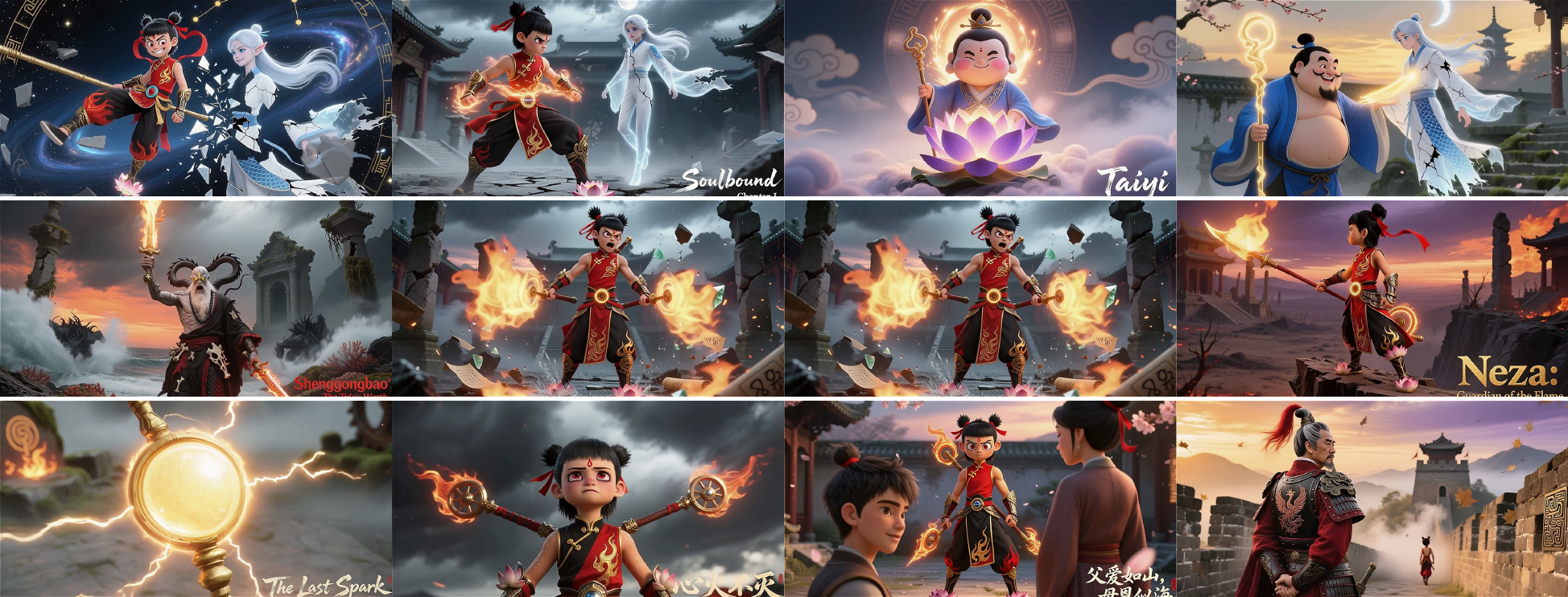}
  \caption{\textbf{Additional qualitative case: mythic fantasy narrative.}
  The generated story maintains stylized visual design and coherent narrative development across ritual, conflict, resolve, and farewell scenes.}
  \label{fig:qual4}
\end{figure*}
\subsection{Character Stills Prompt}

To build persistent character references, DramaAgent first extracts the main characters and their visual attributes from the expanded story. It then generates a canonical visual reference (\textit{Character Still}) for each character.
A representative prompt template is shown below.
\begin{quote}
\small
\textbf{Input:} Character name and visual attributes (e.g., hairstyle, clothing, facial appearance, age, style cues).

\textbf{Instruction:} Generate a high-quality portrait of the character that preserves the specified identity-defining traits. The portrait should serve as a canonical reference image for subsequent scene generation. Maintain consistent facial structure, hairstyle, costume, and overall artistic style.
\end{quote}

These generated character stills form the persistent character reference set $\mathcal{R}=\{R_c\}$ used throughout scene generation and reflective repair.

\subsection{Reflective Rewrite Prompt}
When the reflective repair module detects a low-quality scene candidate, DramaAgent applies a failure-specific prompt rewriting strategy.
A representative reflective rewrite prompt is shown below.
\begin{quote}
\small
\textbf{Input:} Current scene description, detected failure type, previously accepted scene summary, and character references.

\textbf{Instruction:} Rewrite the scene prompt to correct the detected failure while preserving the original story intent. If the failure is identity inconsistency, strengthen appearance and character-reference constraints. If the failure is semantic omission, explicitly inject missing entities, actions, or environment details. If the failure is temporal discontinuity, add transition information from the previous scene. If the failure is audio-visual mismatch, adjust dialogue timing or visual pacing constraints.
\end{quote}

This rewritten prompt is then used for targeted regeneration rather than full-sequence re-generation.

\subsection{Reflection Thresholds and Repair Settings}
In our experiments, DramaAgent generates 3 candidate clips per scene and performs at most 2 rounds of reflective repair.
The reflective module evaluates candidate clips with four criteria: identity consistency, semantic fidelity, temporal continuity, and audio-visual compatibility.
These criteria are aggregated into the reflection score described in the main text.

The repair process is triggered when the best candidate score falls below the acceptance threshold $\tau$. Repair terminates when either the maximum number of rounds is reached or the score improvement between two successive rounds falls below $\epsilon$. In practice, these settings provide a good balance between output quality and computational cost.

\section{Additional Qualitative Results}
\label{app:qualitative}
This section provides additional qualitative examples beyond the representative cases shown in the main paper.
We include diverse story generations to further illustrate the genre generalization, multi-scene coherence, and character consistency of DramaAgent under long-form narrative generation.

Figures~\ref{fig:qual1}--\ref{fig:qual4} present additional qualitative case studies spanning multiple narrative genres, including animated family adventure, action-crime ensemble, fantasy mystery, and mythic fantasy stories.
These examples complement the main-paper visualizations by showing that DramaAgent is not restricted to a single visual style or story template.
Instead, the framework supports long-form generation across substantially different genres while preserving scene-level progression and recurring character identity.

Across these examples, several consistent patterns can be observed.
First, DramaAgent maintains recognizable character appearance across multiple scene transitions, even when the story changes location, lighting, or emotional tone.
Second, the generated stories exhibit coherent narrative development rather than isolated clip-level plausibility: early frames establish characters and setting, middle frames develop conflict, and later frames move toward confrontation or resolution.
Third, the framework remains effective across diverse visual regimes, including cinematic live-action prompts, stylized animation, fantasy environments, and mythology-inspired compositions.

Figure~\ref{fig:qual1} shows an animated family-oriented adventure with recurring characters and light comedic interactions, highlighting the ability of DramaAgent to preserve expressive identity and warm cross-scene continuity.
Figure~\ref{fig:qual2} presents an action-crime narrative with multi-character coordination, tactical progression, and dynamic urban scenes, illustrating that the framework can support visually complex stories with sustained identity consistency.
Figure~\ref{fig:qual3} demonstrates a fantasy mystery narrative centered on clue discovery, mentor guidance, and magical exploration, showing that DramaAgent can maintain atmospheric consistency and structured narrative progression in dialogue- and world-building-heavy stories.
Figure~\ref{fig:qual4} further presents a mythology-inspired story with ritual restoration, large-scale conflict, personal resolve, and emotional farewell, suggesting that the framework can preserve both stylized world design and long-horizon emotional continuity.

Overall, these qualitative cases support the central claim of this work: by combining structured planning, persistent character conditioning, and reflective refinement, DramaAgent improves long-form story coherence across diverse prompts and visual styles.

\section{Discussion}
DramaAgent takes a control-oriented view of long-form text-to-video-and-audio generation.
Rather than introducing a new backbone, it studies how structured planning, persistent character conditioning, and reflective repair can improve story-level coherence over extended generation horizons.
Our results suggest that this agentic control layer is a practical way to make long-form generation more controllable and more robust across heterogeneous video generators.

At the same time, evaluating long-form audiovisual generation remains an important open problem.
Current automatic metrics mainly capture clip-level visual properties, such as consistency, smoothness, or aesthetic quality, but are still limited in measuring story-level coherence across scenes, long-horizon character persistence, and alignment between generated audio and visual events.
As a result, even when automatic scores improve, they may not fully reflect whether a generated story is perceived as globally coherent by human viewers.

We see two important directions for future work.
First, the field would benefit from richer long-form multimodal benchmarks that jointly evaluate narrative progression, identity consistency, temporal continuity, and audio-visual alignment over extended sequences.
Second, more reliable automatic evaluators are needed for story-level assessment, especially evaluators that can reason jointly over text, video, and audio rather than scoring each modality in isolation.
Such advances would make it easier to measure progress on coherent long-form generation and to build stronger feedback loops for automatic repair and refinement.

\end{document}